# Deep Learning to Predict Late-Onset Breast Cancer Metastasis: the Single Hyperparameter Grid Search (SHGS) Strategy for Meta Tuning Concerning Deep Feed-forward Neural Network

Yijun Zhou, Om Arora-Jain, Xia Jiang

Department of Biomedical Informatics
University of Pittsburgh
Pittsburgh, PA

**Contact:**       Xia Jiang

**Email:**         xij6@pitt.edu

**Phone:**         412-648-9310

Yijun Zhou:  yiz209@pitt.edu
Om Arora-Jain: omarorajain@gmail.com

## ABSTRACT

**Background**

While machine learning has advanced in medicine, its widespread use in clinical applications, especially in predicting breast cancer metastasis, is still limited. We have been dedicated to constructing a DFNN model to predict breast cancer metastasis n years in advance. However, the challenge lies in efficiently identifying optimal hyperparameter values through grid search, given the constraints of time and resources. Issues such as the infinite possibilities for continuous hyperparameters like l1 and l2, as well as the time-consuming and costly process, further complicate the task.

**Methods**

To address these challenges, we developed Single Hyperparameter Grid Search (SHGS) strategy, serving as a preselection method before grid search. Our experiments with SHGS applied to DFNN models for breast cancer metastasis prediction focus on analyzing eight target hyperparameters: epochs, batch size, dropout, L1, L2, learning rate, decay, and momentum.

**Results**

We created three figures, each depicting the experiment results obtained from three LSM-I-10+ year datasets. These figures illustrate the relationship between model performance and the target hyperparameter values. For each hyperparameter, we analyzed whether changes in this hyperparameter would affect model performance, examined if there were specific patterns, and explored how to choose values for the particular hyperparameter.

**Conclusions**

Our experimental findings reveal that the optimal value of a hyperparameter is not only dependent on the dataset but is also significantly influenced by the settings of other hyperparameters. Additionally, our experiments suggested some reduced range of values for a target hyperparameter, which may be helpful for "low budget" grid search. This approach serves as a prior experience and foundation for subsequent use of grid search to enhance model performance.

# INTRODUCTION

Machine learning (ML) has always been an important research topic of AI, and we have seen successful cases of using AI-based learning techniques such as deep learning to conduct various tasks and solve a wide range of problems in the biomedical domain. Machine learning and deep learning methods have been used in applications such as predicting drug-drug interactions using real world data collected via large-scale projects [1], modeling miRNA-mRNA interactions that cause phenotypic abnormality in breast cancer patients [2], prioritizing disease-causing gene using sequence-based features candidates[3–5] and predicting ubiquitination sites using physicochemical properties of protein sequences data [6,7]. Various learning methods have also been developed and applied to cancer related prediction such as predicting breast cancer local recurrence using language processing [8], identifying risk factors of prostate cancer recurrence using high-dimensional gene and clinical data[9], and predicting relapse in childhood acute lymphoblastic leukemia (ALL) [10].

An *Artificial Neural Network* (*ANN*) is a machine learning framework, which is designed to recognize patterns using a model loosely resembling the human brain [11,12]. *Deep Neural Networks* (*DNNs*), called *deep learning*, refers to the use of neural networks composed of more than one hidden layer [13–15]. The DNN has obtained significant success in commercial applications such as voice and pattern recognition, computer vision, and image processing [16–19]. However, its power has not been fully explored or demonstrated in clinical applications, such as the prediction of breast cancer metastasis, which is in part due to modeling challenges resulted from the sheer magnitude of the number of variables involved in these problems [20].

We have previously developed *deep feedforward neural network* (DFNN) models that is able to predict n-year breast cancer metastasis [21]. The DFNN models we developed are fully connected neural networks that do not contain cycles. Figure 1 illustrates the structure and the inner connections of a DFNN that we have developed. It is a four-layer neural network that contains one input layer, two hidden layers, and one output layer. The input nodes to this neural network represent the clinical features contained in the input patient data, which we also refer to as predictors, and the output layer contains two nodes representing the binary status of n-year breast cancer metastasis, which also referred to as the target variable in this context.

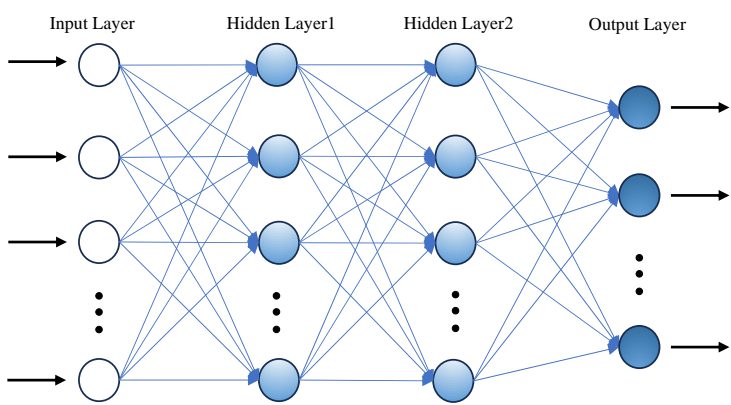

**Figure 1**. The structure of an example DFNN model that contains two hidden layers

One of the challenges of deep learning is that there are a large set of hyperparameters which must be tuned to obtain good prediction models [22]. In machine learning and deep learning, hyperparameters are the variables that determine the model's architecture and directly influence the training process and output model performance [23]. Hyperparameters are predetermined prior to the initiation of the training cycle, and they remain constant, withstanding any learning or modification throughout the progression of the training process [22]. Tuning hyperparameters for a given dataset is a repetitive process of identifying an optimal set of hyperparameter values that produces good prediction results. We call one value assignment to the set of hyperparameters of the DFNN models a *hyperparameter setting* [21]. Grid search is a systematic way of hyperparameter tuning, in which each of a predetermined set of hyperparameter settings will be used to train models and all trained models will be compared to identify the best performing model [21]. We previously conducted grid searches to optimize the prediction performance of our DFNN models [24]. During such a grid search, DFNN models were trained and tested corresponding to each of the preselected hyperparameter settings one at time and the average training and testing results per setting were saved.

The grid searches take as an input a finite number of preselected values for each of the tunable hyperparameters. Some of the hyperparameters of the DFNN models are continuous variables and therefore can assume an infinity number of different values. Preselecting a suitable and finite set of values for such a hyperparameter can be challenging. Depending on the input number of hyperparameter settings, a grid search can be very time consuming and costly. With a low-budget grid search for which available computation time and resource is limited, the number of values that each hyperparameter can take is very small, often making it a difficult task to preselect the input hyperparameter settings, especially for a hyperparameter that has a very large or even an infinity number of possible values [24].

During the preselection of the input hyperparameter settings, we often face unanswered or under-answered questions such as 1). Will changing values of a particular hyperparameter truly has a significant effect on model performance? 2). How does model performance change as the value of a hyperparameter increases? are there particular patterns? 3). If there are performance patterns associated with a hyperparameter, would they be dataset dependent? 4). When selecting a hyperparameter value, should we be concentrating on lower values, mid values, or higher values? 5). With a low-budget grid search, what would be a rule of thumb for selecting a promising set of values for a hyperparameter that can take a very large number of values?

In this paper, we introduce the single hyperparameter grid search (SHGS) strategy, which can be used prior to a grid search to help conduct the challenging task of preselecting the input hyperparameter settings for the grid search. We describe this strategy in detail in the Method section below. We conducted experiments using the SHGS with our DFNN models for predicting late onset breast cancer metastasis. In this context, we refer to a breast cancer metastasis that occurred after 10 plus years of the initial breast cancer treatment as a late onset breast cancer metastasis. We identified eight tunable hyperparameters that have a large or an infinity number of values including epochs, batch size, dropout, L1, L2, learning rate, decay, or momentum, which are referred to as target hyperparameters in this study. Table 1 shows detailed information about the tunable variables in the DFNN models including these target hyperparameters. We ran SHGS experiments multiple times for each of the eight target hyperparameter. Through these experiments, we hope to gain insights about preselection of the input hyperparameter settings required by a grid search, which can be useful for researchers who are interested in conducting predictions using similar types of datasets and techniques.

## METHODS

**The Single Hyperparameter Grid Search (SHGS) strategy**

The purpose of running a SHGS is to look for a promising but reduced range of values for a particular hyperparameter, which we call the target hyperparameter in the SHGS, to assist the task of pre-selecting input values for the subsequent grid search. In a SHGS, we will first give a large range of values for the target hyperparameters that has an infinity number of possible values such as L1 and L2, while all the other hyperparameters each will take only one preselected value. So, a SHGS can be considered as a special type of grid search in which the number of hyperparameter settings used for model training during the grid search is equal to the number of values of the target hyperparameter. In this study, we identified 8 hyperparameters that can take an infinity number of values and treat them as our target hyperparameters. These 8 hyperparameter are epochs, batch size, learning rate, dropout rate, momentum, decay, L1 weight decay, and L2 weight decay. Table 1 shows a description of the 8 target hyperparameters and their values that we studied via our SHGS experiments. In these experiments, to avoid bias, the fixed value used by each of the non-target hyperparameters is pre-picked randomly from the pool of all values considered for each of these hyperparameter following a uniform distribution. We call one value assignment of these non-target hyperparameters a background hyperparameter setting. Table 1 also shows all values of the non-target hyperparameters that are considered in this research. For each of the target hyperparameter, we repeat the SHGS experiment 10 times and each time use a different randomly selected background hyperparameter setting from all possible settings, to see how a different background setting can affect the results. We used the Python programming language along with the TensorFlow and Keras libraries to implement SHGS, which is now available at [https://pypi.org/project/SHGS].

**Table 1**. Hyperparameters and their values used in SHGS experiments

| Hyperparameters | Description | Values Used in SHGS Experiments | Number of values studied per SHGS experiment (target hyperparameter only) |
|---|---|---|---|
| Epochs | Number of times model is trained by the full training dataset | 5 ~ 1001, step size 3 | 333 |
| Batch size | Number of samples that are processed together in a single forward and backward pass during training | 1 to the # of datapoints in a dataset | 730 in LSM-10Years-I dataset 446 in LSM-12Years-I dataset 300 in LSM-15Years-I dataset |
| Learning rate | Control the learning and parameter update speed during optimization | 0.001 ~ 0.3, Step size 0.001 | 300 |
| Dropout rate | Mitigate overfitting and training time by randomly ignoring nodes | 0 ~ 0.9, Step size 0.01 | 91 |
| Momentum | Speed up optimization by incorporating historical gradients into parameter updates. Momentum is exclusively applicable within the SGD optimizer. | 0.1 ~ 0.9, Step size 0.01 | 81 |
| Decay | Iterative decay of the learning rate by applying a decreasing factor at each epoch | 0 ~ 0.3, Step size 0.001 | 301 |
| L1 weight | Control the strength of L1 regularization in model training | 0 ~ 0.3, Step size 0.001 | 301 |
| L2 weight | Control the strength of L2 regularization in model training | 0 ~ 0.3, Step size 0.001 | 301 |

| # of Hidden Layers | The depth of a DNN model | 1,2,3,4 | Non-target hyperparameter |
|---|---|---|---|
| # of Hidden Nodes | Number of neurons in a hidden layer | All integers from 1 to the number of datapoints of each dataset | Non-target hyperparameter |
| Optimizer | Optimizes model parameters during training process towards minimizing the loss | SGD, Adam, Adagrad, Nadam, Adamax | Non-target hyperparameter |
| Weight Initializer | A technique employed to assign initial values to the weights of the connections between neurons in a neural network | Constant, Glorot_normal, Glorot_uniform, He_normal, He_uniform | Non-target hyperparameter |
| Input Layer Activation Function | A function applied to the input data of a neural network's input layer | Relu, Sigmoid, Softmax, Tanh | Non-target hyperparameter |
| Hidden Layer Activation Function | A function applied to the output of a hidden layer in a neural network | Relu, Sigmoid, Softmax, Tanh | Non-target hyperparameter |
| Output Layer Activation Function | A function applied to the output of a neural network's output layer, producing the final prediction of the network. | Sigmoid | Non-target hyperparameter |
| Loss Function | A method of evaluating the dissimilarity between the predicted output of a model and the actual values | Binary_crossentropy | Non-target hyperparameter |

**Experiments**

Figure 2 is a flow chart that demonstrates the experiments we did involving using SHGS. The whole procedure consists of three inputs: number of iterations, target hyperparameter, and hyperparameter ranges for all hyperparameters. The number of iterations indicates the number of times the SHGS algorithm is executed based on the input target hyperparameter. With each iteration of the SHGS calls, it selects a value randomly for each of the non-target hyperparameters. These randomly selected values for all non-target hyperparameters are combined into a background hyperparameter setting. We systematically explore each value within the specified range of values for a target hyperparameter, forming a target hyperparameter setting in conjunction with the background setting. A call to SHGS will initialize the target hyperparameter to progress sequentially through the range of the target hyperparameter and conducts 5-fold cross validation (see details below) to train models based on each of the different target hyperparameter settings. It also tests the output model with an independent test dataset that was set aside and saves the results.

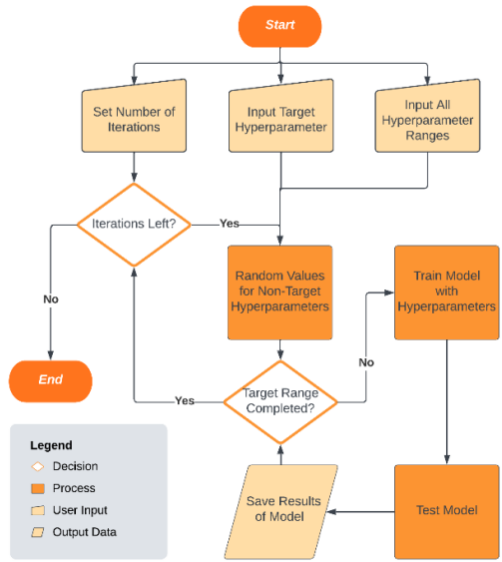

**Figure 2.** A flow chart illustrating the experiments we did using SHGS.

**The Evaluation of Model Performance Using 5-fold Cross Validation**

For a given binary diagnostic test, a receiver operator characteristic (ROC) curve plots the true positive rate against the false positive rate for all possible cutoff values [25]. The area under a ROC curve (AUC) measures the discrimination performance of a model [26]. We followed a 5-fold cross validation (CV) procedure to train and evaluate models under each hyperparameter setting in a SHGS. The dataset was first split into two subsets: an 80% training dataset for optimizing the model's performance and a separate 20% independent test dataset to evaluate its generalization ability. The entire training dataset was then divided evenly into 5 sub-datasets. The division was mostly conducted randomly and, in the meantime, we made sure that each sub-dataset had approximately 20% of the positive cases and 20% of the negative cases to guarantee an overall balanced distribution of the two. For each target hyperparameter setting, we conducted both the training and testing five times. At each time a model is learned from a different set of 4 sub-datasets combined, and then it is tested using the remaining sub-datasets as the validation set. The average training and testing AUC across all five times, which are called mean_train_auc and mean_test_auc, were reported via the grid search package. Additionally, to evaluate the model's performance completely independently, we used the 20% set-aside test dataset to test the output model refitted using the entire training dataset under each target hyperparameter setting to obtain a test_auc. We used this procedure for all datasets involved in this research.

**Datasets**

In this research, we used three datasets LSM-I-10Year, LSM-I-12Year, and LSM-I-15Year [27,28]. The letter I stands for interactions. We previously developed the MBIL method to retrieve interactive features that has a direct influence on a target feature such as a disease outcome [27]. Such a target feature is often called the class feature in machine learning. In this context, the non-target features are also referred to as predictors. LSM-nYear datasets were developed in previous studies [27,28]. In this study we applied the MBIL method to obtain the sets of interactive features from the LSM Datasets, then we retrieved the LSM-I-10Year, LSM-I-12Year, and LSM-I-15Year datasets from the corresponding LSM-nYear datasets based on these interactive features. Each of the three datasets contains the target feature "metastasis". Table 2 below shows the counts of the cases and predictors included in the three datasets. A description of these predictors is included in the Supplementary Tables S1-S3 and a previously published paper [29].

Table 2. Case and predictor count of the three LSM-I-10[+] year datasets.

|  | Total # of cases | # Positive cases | # Negative cases | #Number of Predictors |
|---|---|---|---|---|
| LSM-I-10year | 1827 | 572 | 1255 | 18 |
| LSM-I-12year | 1115 | 588 | 527 | 20 |
| LSM-I-15year | 751 | 608 | 143 | 17 |

## RESULTS

Figures 3, 4, and 5 are the results obtained for the LSM-I-10Year, LSM-I-12Year, and LSM-I-15Year datasets, each respectively. Each figure consists of eight panels of scatter plots, one for each of the eight target hyperparameters. Within each panel are ten individual scatter plots, each showing the results of a specific SHGS experiment conducted using the corresponding dataset for the corresponding target hyperparameter. Such a scatter plot demonstrates how model performance, as measured by test_auc, changes when the values of a target hyperparameter increases during a SHGS experiment.

Based on the results showed in Figure 3-5, the scatter plots of the 10 different experiments for the same target hyperparameter can be quite different. This is perhaps because each experiment used a different background setting that was randomly selected by the SHGS scheme. Note that with certain hyperparameter configurations that were randomly generated via grid search, the prediction performance can be very poor, indicated by the consistently low test_AUC values (approximately 0.5), as seen in figures such as Figure 3(a) experiments 7 and 8. An AUC value of 0.5 indicates that the model is unable to distinguish between positive and negative class points, performing no better than random guessing [30].

As to epochs (Figure 3(a) through 5(a)), the highest test_auc values for the 10-year, 12-year, and 15-year data are 0.683 (Figure 3(a)-5), 0.727 (Figure 4(a)-8), and 0.852 (Figure 5(a)-2), each respectively. We notice that optimal test_auc values are often reached at low epochs (examples: Figure 3(a)-5 and Figure 5(a)-2). Besides, we have seen three different patterns: 1) test_auc decreases once the number of epochs surpasses a specific threshold (examples: Figure 3(a)-2 and Figure 5(a)-4), 2) test_auc plateaus after passing certain point (example, Figure 4(a)-8) or throughout (example, Figure F5(a)-1), and 3) test_auc steadily goes up as number of epochs passes certain point (examples: Figure 4(a)-1, and Figure 5(a)-10. We also notice that in some cases test_auc has high variance and fluctuates rapidly while the number of epochs increases (example, Figure 4(a)-6), while it has very low variance in some other cases (example, Figure 5(a)-1).

In the batch size experiments (Figure 3(b) through 5(b)), the best test_auc values are 0.7 (Figure 3(b)-6) for 10-year, 0.738 (Figure 4(b)-8) for 12-year, and 0.856 (Figure 5(b)-10) for 15-year. There tend to be a short warm up period (batch size< 50) when performance is unstable but reaches a peak quickly, and in most experiments over all three datasets performance reaches the peak before batch size increases to 1/5 of its highest value. Overall, we see three patterns after performance reaches its peak: 1) a slight negative correlation between performance and batch size (Figure 3(b)-7, 4(b)-9, 5(b)-4, and 5(b)-5; 3) a sharp dip when batch size reaches about half its highest value, but trend remains stable at either side of the sharp dipping point (Figure 3(b)-5, 3(b)-9, Figure 4(b)-3 and 4(b)-7, and Figure 5(b)-6; and 3) trend remains constant, with or without a big variance, and this is seen in most cases, indicating that larger batch size tends to have limited influence on performance improvement.

Figure 3-5(c) shows the patterns of dropout rate, the highest scores are 0.77 (10 Year), 0.726 (12 Year), and 0.886 (15 Year). Similar to batch size, in most cases, the performance trend remains constant or gets worse after it reaches a quick peak. The performance tends to become very poor after the dropout passes 0.5, with some exceptions (examples: Figure 3(c)-5, 4(c)-1, 5(c)-7).

Regarding the L1 and L2 regularization parameters, denoted as L1 and L2 respectively in subsequent discussions, the best test_auc values for L1 in Figure 3(d)-5(d) are 0.689, 0.734, and 0.872, and the best test_auc results for L2 in Figure 3(e)-5(e) are 0.706, 0.734, and 0.86. The results of L1 reveal no increasing trend in performance as measured by test_auc when L1 values increase in any of the 10 experiments. Based on these results, using a small L1 value (<0.03) in grid search can be sufficient to obtain the best-performed model. L2 results demonstrate similar trends in most cases with some exceptions, in which it requires a larger L2 value to reach the performance peak in experiments 3(e)-2, 3 (3)-10, 4(e)-2, 4(e)-4, and 5(e)-5), but none of these experiments gives the best test_auc values.

The highest test_auc scores for learning rate (Figure 3(f) through 5(f)), are 0.734 (Figure 3(f)-6), 0.762 (Figure 4(f)-4), and 0.832 (Figure 5(f)-5), each respectively. The best test_auc values in most experiments were obtained when learning rate is below 0.03. After reaching the peak, the performance tends to stabilize (example, Figure 4(f)-2), or alternatively, it exhibits a significant decline with increased fluctuations after surpassing a certain point (example, Figure 5(f)-10).

The best test_auc values achieved for decay are 0.706 (Figure 3(g)-10 for 10 year), 0.733 (Figure 4(g)-2 for 12 year) and 0.87 (Figure 5(g)-4 for 15 year). All experiments that achieved the best test_auc values reach the best performance almost immediately and the increasing value of decay has no significant impact on performance. Certain patterns are seen for other experiments such as the 8th experiment of Figure 4(g), where the test_auc value initially rises as the decay increases but drops abruptly at a point (around 0.03) and remains suboptimal. Overall, the performance reaches the peak when decay assumes a small value (below 0.03), then the trend either maintains constant or goes down as decays increase, with exceptions as seen in Figure 5(g)-2 and 5(g)-6, where it takes longer for the performance to reach the peak.

For momentum, the best test_auc values are 0.675 (Figure 3(h)-3 for 10 year), 0.72 in Figure 4(h)-6 for 12 year), and 0.835 (Figure 5(h)-5 for 15 year). Considering all three datasets, performance trend mostly demonstrates two patterns: stays constant or goes up after momentum at least passes the half point, but the well-performed models tend to appear with the second pattern. For examples, for the 10-year data, the performance reaches the peak after momentum passes 0.5 in the experiment in which we identify the best model (Figure 3(h)-3), and for the 15-year data, we obtain the best model after momentum passes 0.7. For the 12-year dataset, the best test_auc is obtained when momentum is below 0.3 and the trend after the peak test_auc is going down as an exception, but we obtain near-best models in other experiments when the momentum passes 0.5 (Figure 4(h)-1) and when it passes 0.7 (Figure 4(h)-2).

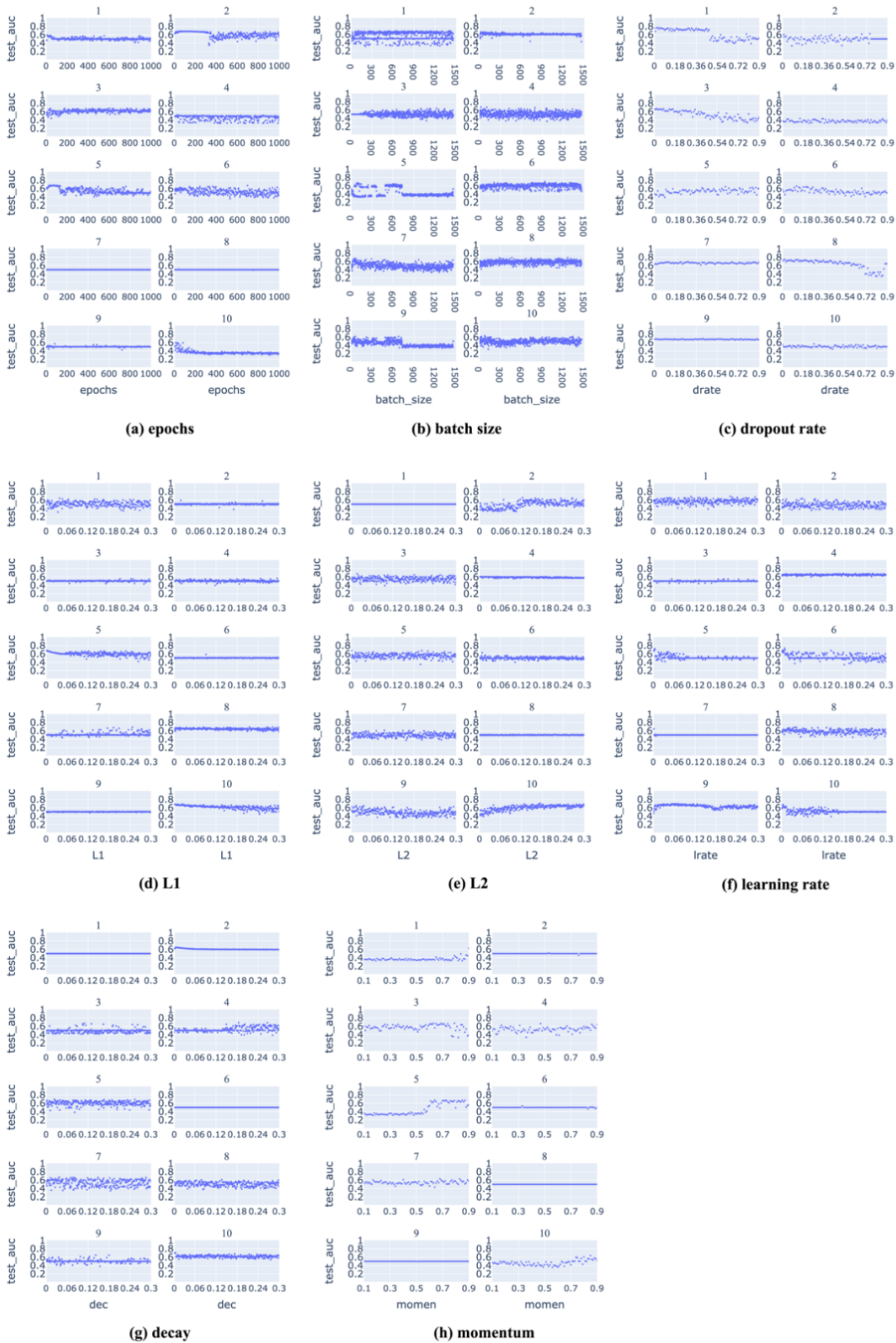

**Figure 3.** Scatter plots: test_auc vs a target hyperparameter concerning LSM-I-10Year

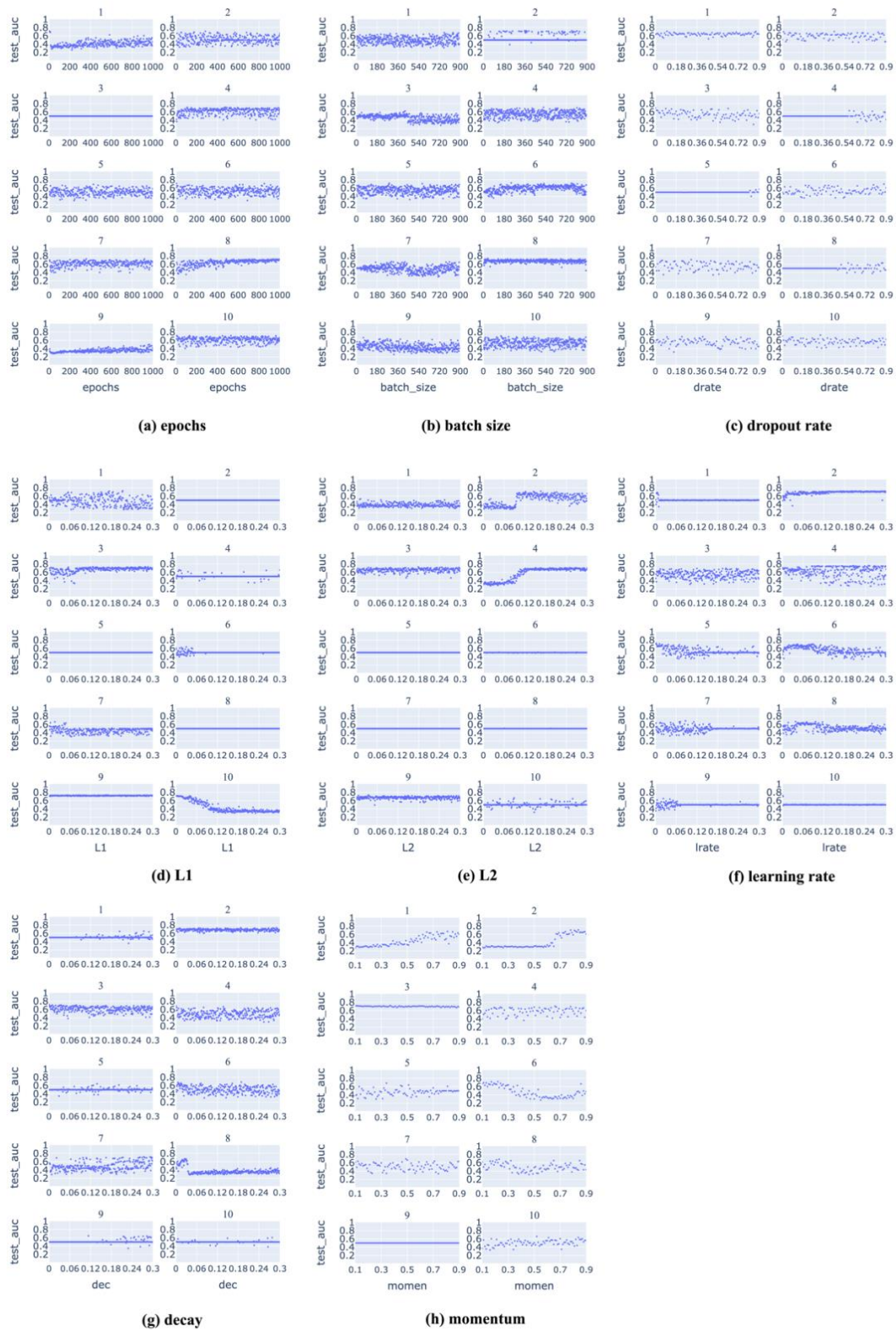

**Figure 4:** Scatter plots: test_auc vs. a target hyperparameter concerning LSM-I-12Year

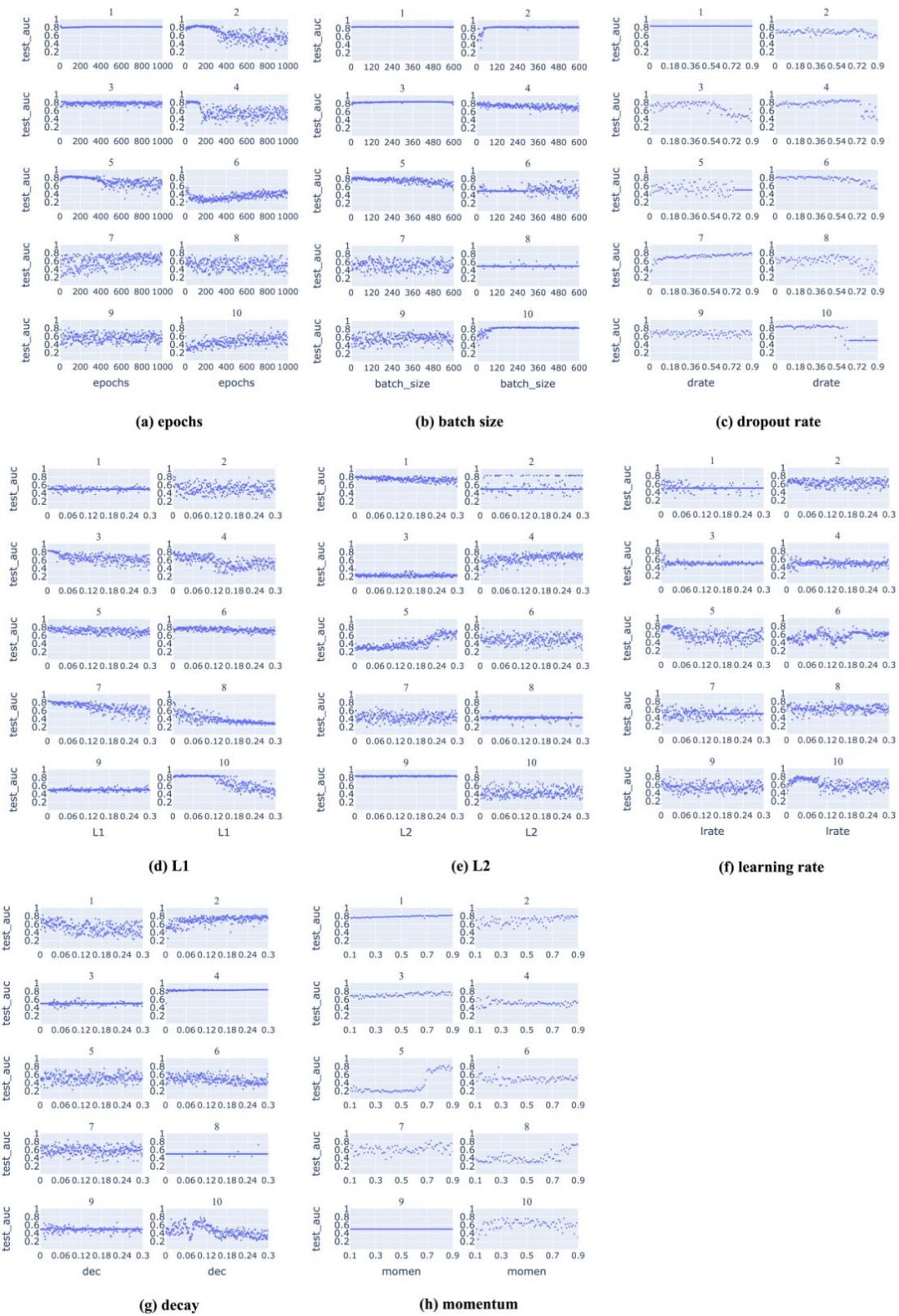

**Figure 5.** Scatter plots: test_auc vs. a target hyperparameter concerning LSM-I-15Year

**Table 3.** Number of hyperparameter settings and running time per dataset in SHGS experiments

| Dataset | Number of Hyperparameter Settings | Unit Running Time Per Hyperparameter Setting | Total Running Time Per Dataset |
|---|---|---|---|
| LSM-I-10Year | 24380 | 93.875s | 635.74h |
| LSM-I-12Year | 21540 | 71.454s | 427.532h |
| LSM-I-15Year | 20080 | 76.849s | 428.646h |

Table 3 displays the number of hyperparameter settings and corresponding running times for the SHGS experiments conducted using the three LSM-I datasets. Table 4 below represents the total experimental time for each of the 8 target hyperparameters studied across the same datasets. Surprisingly, the experiments targeting L2 as the hyperparameter exhibit the longest duration among all three datasets.

**Table 4:** Running time (in hours) per target hyperparameter

| Target Hyperparameter | Total Running Time | Running time per dataset | | |
|---|---|---|---|---|
| | | LSM-I-10Year | LSM-I-12Year | LSM-I-15Year |
| Epochs | 178.003 | 113.047 | 29.568 | 35.388 |
| Batch size | 145.69 | 77.542 | 59.293 | 8.855 |
| L1 | 170.957 | 115.336 | 47.719 | 7.903 |
| L2 | 466.878 | 103.453 | 161.839 | 201.586 |
| Dropout rate | 102.28 | 51.569 | 10.393 | 40.318 |
| Learning rate | 125.213 | 23.894 | 64.082 | 37.237 |
| Momentum | 107.467 | 85.205 | 5.818 | 16.444 |
| Decay | 195.429 | 65.694 | 48.82 | 80.915 |

We also study how running time changes along with each of the hyperparameters, but as shown in Supplementary Figures1-3, there are no apparent correlations between running time and a target hyperparameter in most cases. There are a few notable correlations between the running time and a target hyperparameters such as batch size, epochs, or L1 as shown in Figure 6 below. Based on Figure 6(a), running time decreases quickly as the value of batch size increases till it reaches a turning point, from which the running time either remains unchanged or decreases very slowly and slightly. The turning point often occurs when the batch size is very small, mostly below 50. Figure 6(b) shows there is often an apparent positive correlation between the running time and epochs. Figure 6(c) demonstrates occasionally there is a positively correction between the running time and L1, but in most experiments, increasing the value of L1 does not affect the running time much.

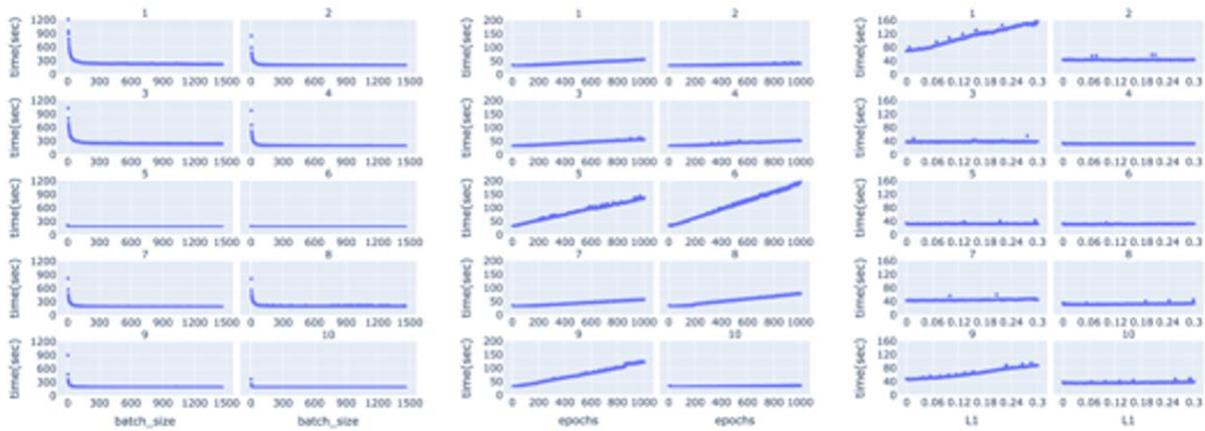

(a) batch_size with LSM-I-10Year  (b) epochs with LSM-I-12Year  (c) L1 on LSM-I-15Year
**Figure 6.** Selected scatter plots: running time vs a target hyperparameter

## DISCUSSION

Previous studies reported that best model performance can be obtained at very high epochs [31], or at very low epochs values such as below 10 [32]. One possible explanation is that these experiments were done using different datasets, and a previous study reveals that there is no one-value-fits-all optimal value for epochs, and the value of epochs heavily depends on the dataset [33]. Our results demonstrate that epochs are not only dataset dependent, but also perhaps depending on the background hyperparameter setting. Recall that we refer to a background hyperparameter setting or a background setting as one unique value assignment of all non-target hyperparameters. For example, the 10 scatter plots in Figure 5(a) show that the effects of epochs can be quite different at different background settings for the same dataset. Optimal models are only obtained at low epochs with setting 2, 4, and 5, but at setting 1 and 3, model performance was not deteriorated as epochs goes high and we obtain well performed models when epochs reach 1000, the highest value we tested. For all three datasets, we see optimal models at low epochs (<100) as seen in Figure 3(a)-5, 4(a)-7, and 5(a)-2. This is consistent with a previous finding reported in [34], which states deep learning often converge within low epochs. Besides, our results showed (Figure 6) running time is often correlated with number of epochs, so using high epochs could significantly increase computation time, which renders it a lessor choice especially when time is limited to run a grid search.

The optimal batch size remains a subject of ongoing investigation, and no universally agreed-upon answer has been established. Several studies, such as [35] have observed that smaller batch sizes tend to yield improved results. Conversely, another study [36], contends that larger batch sizes lead to higher testing accuracy. This inconsistency may again be due to the different dataset and different background hyperparameter configures used in these studies. Based on our results, both small batch size (for example Figure 4(b)-8) and large batch size (example Figure 5(b)-2) can lead to optimal models, and at certain background hyperparameter settings model performance remains optimal regardless we use a small or a large batch size (examples Figure 5(b)-1 and 5(b)-2). So like epochs, batch size seems to be both dataset and background setting dependent. Our results also show that performance trend most often declines after quickly reaching a peak at a small batch size that is about below 1/5 of its maximum value. This observation supports that large-batch training methods tend to converge towards sharp minimizers, resulting in reduced generalization ability previously reported in[34,37]. It also suggests that choosing batch size values that are below 1/5 of the maximum value might be a good guideline for grid search, especially when computation time is limited and the number of batch size values that can be selected is very small due to the time constraint [38].

Finding a dropout rate that works perfectly for all datasets and situations is difficult, as it is influenced by various factors [35,36,39,40]. The dataset size seems to be a key factor. Research indicate that smaller dropout rates are beneficial for small datasets [41], and remain effective for large datasets. Our results showed that the performance trend mostly remains constant or goes down after it reaches the peak, which often occurs before dropout reaches 0.5. This substantiated that a dropout rate of p=0.5 works well for various networks as suggested in [20]. Combining our experiment results and the previous findings, using dropout 0.5 or a few values that are less or equal to 0.5 might be a good initial choice for a low-budget grid search.

L1 regularization is suggested to be more effective than L2 regularization and dropout[42–44]. [42] suggested that a good value of L1 value is 0.001 compared to 0.0001, 0.01, and 0.1 that was tested. According to [43], if L1 is too high, the model is too simple, leading to underfitting, while it's too low, the model becomes too complex and may overfit. Based on our experiment results, regardless of the dataset, the result is consistently better when using a smaller L1 value compared to a larger L1 value, and the optimal test_auc values are consistently achieved when L1 was less than 0.03. The variations in test_auc concerning L1 values within a narrower range (0-0.03) are worth investigating. L2 results demonstrate similar trends in most cases with some exceptions, in which it requires a L2 value that is greater than 0.03 to reach the performance peak in experiments 3(e)-2, 3 (3)-10, 4(e)-2, 4(e)-4, and 5(e)-5), but none of these experiments give the best test_auc values. These results suggest that for a low-budget grid search, using a narrow range of L1 and L2 values such as 0 to 0.03, is a good initial choice.

Selecting a range of learning rate values suitable to a grid search can also be challenging due to the uncertainty involved. The learning rate should not only be sufficiently large so that the learning process does not take too much time to complete but also be small enough to not overshoot the optimal convergence or cause large oscillations [45,46]. Based on our experiment result, the optimal learning rate value is not only depending on dataset but also depending on the background setting of an experiment. In each experiment, we tested a large range of learning rate values (0 to 0.3), but across all three datasets, the best test_auc values most often occur before learning rate reaches 0.03. We also notice that once the test_auc reaches a peak, it rarely goes up as learning rate further increases. These experiment results suggest a suitable range of learning rate values for a low budget grid search is below 0.03.

Researchers currently tend to determine the values of learning rate decay (referred to as decay in this context) based on empirical considerations, including 0.1, 0.01, 0.001, among others [44]. Due to the relationship between decay and learning rate, the magnitude of the learning rate is a crucial factor to consider when determining the value of decay [44]. In our study, all experiments that achieve the best test_auc values reach the best performance almost immediately (at a decay close to 0) and the further increasing value of decay has no significant impact on the performance.  For other experiments that did not give the best test_auc values, we notice that the performance reaches the peak mostly when decay assumes a small value that is below 0.03, then the trend either maintains constant or goes down as decays increase. Overall combining with textbook decay values and our experiment results, we suggest choosing decay values from 0 to 0.003 (or 1/10 of the upper bound learning rate) for a low-budget grid search and perhaps using a larger range such as from 0 to 0.03 if budget permits.

Momentum directly influences optimization during model training and expedites convergence by incorporating historical gradients [47]. Previous research suggests that values around 0.9 seem to be good values [43,44]. Some papers, however, claim that the relationship between the values of momentum and learning rate have more of an effect than the individual values themselves [48]. In our experiments, we tested 81 momentum values ranging from 0.1 to 0.9, to be comprehensive. Our results show that optimal models most often appear when momentum is greater than 0.5, and two of the three best test_auc values are obtained when momentum is greater than 0.7. However, we did notice that occasionally an optimal model can be obtained at a low momentum value. For example, the best test_auc for LSM-I-12Year dataset is obtained when momentum is less than 0.3. Overall, our experiments suggest a good initial range of values for momentum is 0.7 to 0.9. If budget permits, a range of low momentum values such as 0.01 to 0.03 can also be included in a grid search.

# CONCLUSIONS

Our experiments results demonstrate that an optimal value of a hyperparameter is not only dataset dependent, but also affected significantly by the value assignment of the other hyperparameters, which we call a background hyperparameter setting. This may indicate that hyperparameters can interact and therefore affect model performance jointly. All eight hyperparameters can lead to optimal models at small values after a short warm up period, and for most of them model performance trend maintains constant or declines after the peak except for epochs, batch size, and momentum. Based on our results, depending on a background hyperparameter configuration, both low and high values of epochs, batch size, or momentum can lead to optimal models, which helps explain the seemingly conflicting findings reported by previous studies. As to providing guidance to grid search, especially when computation time or resource is limited, our results suggest the following reduced range of values: epochs below 100, batch size below 1/5 of the size of datapoints, dropout below 0.5, momentum between 0.7 and 0.9, decay below 0.003, and learn rate, L1, and L2 each below 0.03. Finally, since the optimal values are dataset and background setting dependent, conducting some SHGS experiments can be very helpful to hyperparameter configuration prior to the main grid search.

# DECLARATIONS

**Ethics approval and consent to participate**

The study was approved by University of Pittsburgh Institutional Review Board (IRB # 196003) and the U.S. Army Human Research Protection Office (HRPO # E01058.1a).

The need for patient consent was waived by the ethics committees because the data consists only of de-identified data that are publicly available.

**Consent for publication**

Not applicable.

**Availability of data and material**

The data used in this study are available at datadryad.org (DOI 10. 5061/dryad.64964m0).

**Competing interests**

The authors declare that they have no competing interests.

**Funding**

Research reported in this paper was supported by the U.S. Department of Defense through the Breast Cancer Research Program under Award No. W81XWH-19-1-0495 (to XJ). Other than supplying funds, the funding agencies played no role in the research.

**Authors' Contribution**

XJ originated the study and designed the methods. YZ, OAJ, and XJ wrote the first draft of the manuscript. XJ and YZ implemented the methods. YZ conducted experiments. YZ, OAJ, and XJ prepared and analyzed the results. All authors contributed to the preparation and revision of the manuscript. All work was conducted in the University of Pittsburgh.

**Acknowledgements**

OAJ was involved in the work as a trainee in XJ's Artificial Intelligence research lab summer 2023 via UPMC Hillman Cancer Center Academy.

# Supplement

Table S1. A Description of the Variables of the LSM-10Year Dataset

| Variables included | Description | Values |
|---|---|---|
| *ethnicity* | ethnicity of patient | not Hispanic, Hispanic |
| *smoking* | smoking history of patient | ex smoker, non smoker, cigarettes, chewing tobacco, cigar |
| *alcohol usage* | alcohol usage of patient | moderate, no use, use but nos (non otherwise specified), former user, heavy user |
| *family history* | family history of cancer | cancer, no cancer, breast cancer, other cancer, cancer but nos |
| *age_at_diagnosis* | age at diagnosis of the disease | 0-49, 50-69, >69 |
| *TNEG* | triple negative status in terms of patient being ER, PR, and HER2 negative | yes, no |
| *ER* | estrogen receptor expression | neg, pos, low pos |
| *ER_percent* | percent of cell stain pos for ER receptors | 0-20, 20-90, 90-100 |
| *PR* | progesterone receptor expression | neg, pos, low pos |
| *PR_percent* | percent of cell stain pos for PR receptors | 0-20, 20-90, 90-100 |
| *HER2* | HER2 expression | neg, pos |
| *n_tnm_stage* | # of nearby cancerous lymph nodes | 0, 1, 2, 3, 4, X |
| *stage* | composite of size and # positive nodes | 0, 1, 2, 3 |
| *lymph_nodes_positive* | number of positive lymph nodes | 0, 1-8, >8 |
| *histology* | tumor histology | lobular, duct |
| *grade* | grade of disease | 1, 2, 3 |
| *DCIS_level* | type of ductal carcinoma in situ | solid, apocrine, cribriform, dcis, comedo, papillary, micropapillary |
| *surgical_margins* | whether residual tumor | res. tumor, no res. tumor, no primary site surgery |
| *distant recurrence (Metastasis)* | This is the target variable | yes, no |

Table S2. The variables of the LSM-12Year Dataset

| Variables included | Description | Values |
|---|---|---|
| *race* | race of patient | white, black, Asian, American Indian or Alaskan native, native Hawaiian or other Pacific islander |
| *ethnicity* | ethnicity of patient | not Hispanic, Hispanic |
| *family history* | family history of cancer | cancer, no cancer, breast cancer, other cancer, cancer but nos |
| *age_at_diagnosis* | age at diagnosis of the disease | 0-49, 50-69, >69 |
| *menopausal_status* | inferred menopausal status | pre, post |
| *ER* | estrogen receptor expression | neg, pos, low pos |
| *ER_percent* | percent of cell stain pos for ER receptors | 0-20, 20-90, 90-100 |
| *PR* | progesterone receptor expression | neg, pos, low pos |
| *PR_percent* | percent of cell stain pos for PR receptors | 0-20, 20-90, 90-100 |
| *P53* | whether P53 is mutated | neg, pos, low pos |
| *HER2* | HER2 expression | neg, pos |
| *n_tnm_stage* | # of nearby cancerous lymph nodes | 0, 1, 2, 3, 4, X |
| *stage* | composite of size and # positive nodes | 0, 1, 2, 3 |
| *lymph_nodes_positive* | number of positive lymph nodes | 0, 1-8, >8 |
| *histology* | tumor histology | lobular, duct |
| *size* | size of tumor in mm | 0-32, 32-70, >70 |
| *grade* | grade of disease | 1, 2, 3 |
| *invasive* | whether tumor is invasive | yes, no |
| *DCIS_level* | type of ductal carcinoma in situ | solid, apocrine, cribriform, dcis, comedo, papillary, micropapillary |
| *surgical_margins* | whether residual tumor | res. tumor, no res. tumor, no primary site surgery |
| *distant recurrence (Metastasis)* | This is the target variable | yes, no |

**Table S3.** The variables of the LSM-15Year Dataset

| Variables included | Description | Values |
|---|---|---|
| *race* | race of patient | white, black, Asian, American Indian or Alaskan native, native Hawaiian or other Pacific islander |
| *alcohol usage* | alcohol usage of patient | moderate, no use, use but nos (non otherwise specified), former user, heavy user |
| *age_at_diagnosis* | age at diagnosis of the disease | 0-49, 50-69, >69 |
| *menopausal_status* | inferred menopausal status | pre, post |
| *ER* | estrogen receptor expression | neg, pos, low pos |
| *ER_percent* | percent of cell stain pos for ER receptors | 0-20, 20-90, 90-100 |
| *t_tnm_stage* | prime tumor stage in TNM system | 0, 1, 2, 3, 4, IS, 1mic, X |
| *n_tnm_stage* | # of nearby cancerous lymph nodes | 0, 1, 2, 3, 4, X |
| *stage* | composite of size and # positive nodes | 0, 1, 2, 3 |
| *lymph_node_status* | patient had any positive lymph nodes | neg, pos |
| *size* | size of tumor in mm | 0-32, 32-70, >70 |
| *grade* | grade of disease | 1, 2, 3 |
| *histology2* | tumor histology subtypes | IDC, DCIS, ILC, NC |
| *invasive_tumor_location* | where invasive tumor is located | mixed duct and lobular, duct, lobular, none |
| *re_excision* | removal of an additional margin of tissue | yes, no |
| *surgical_margins* | whether residual tumor | res. tumor, no res. tumor, no primary site surgery |
| *histology* | tumor histology | lobular, duct |
| *distant recurrence (Metastasis)* | This is the target variable | yes, no |

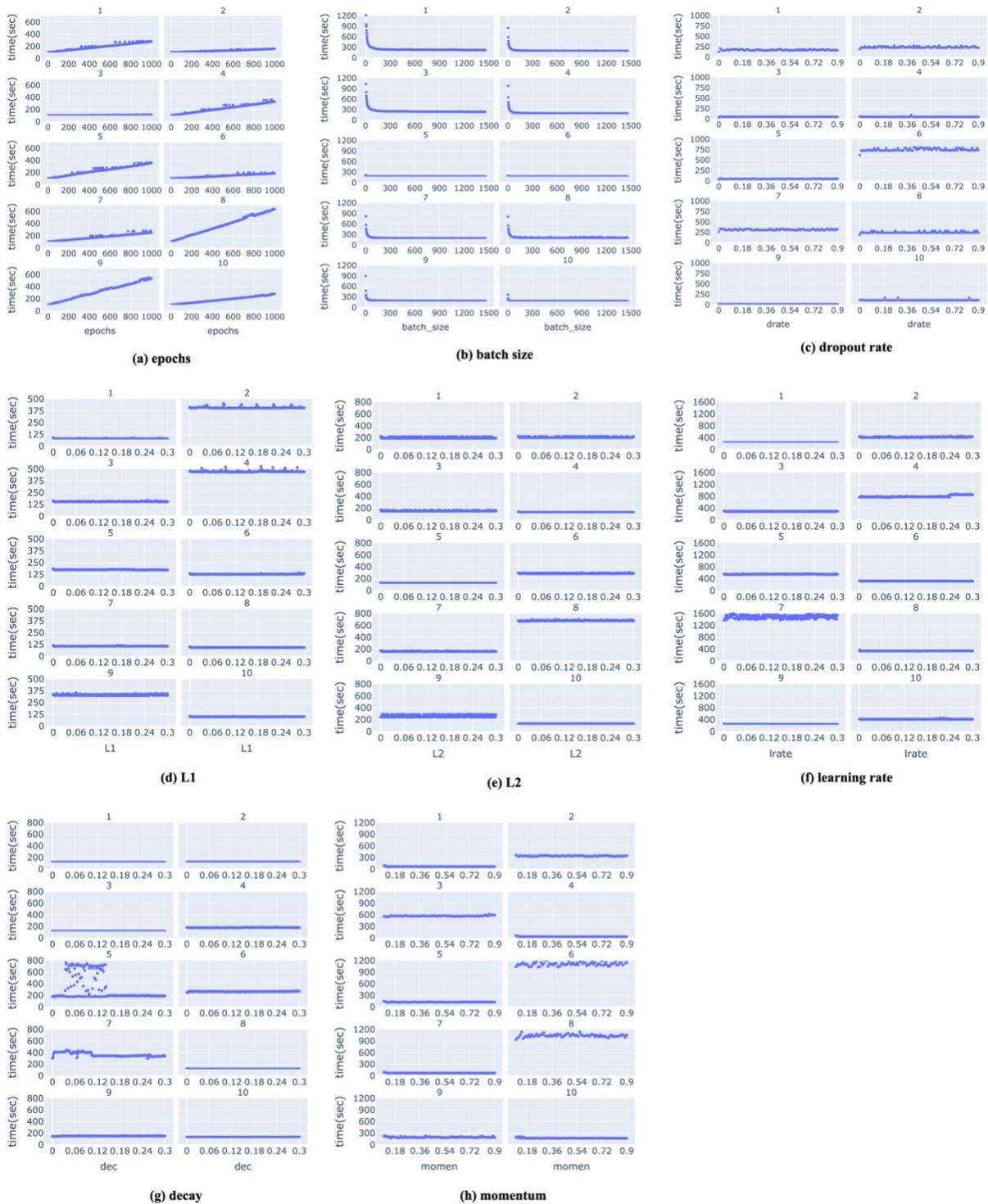

**Figure S1.** Scatter plots: running time vs. the values taken by the target hyperparameters concerning LSM-I-10Year

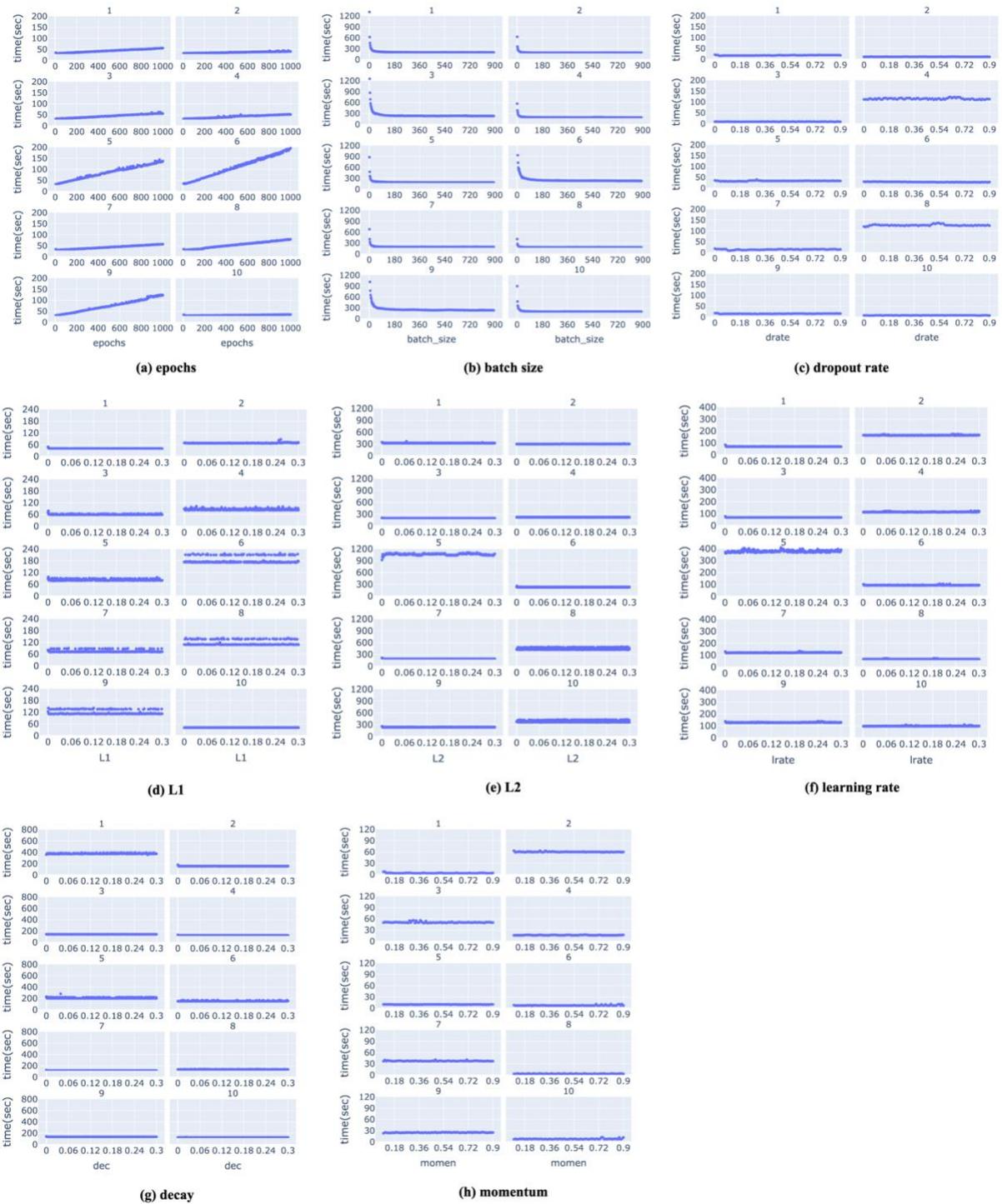

**Figure S2.** Scatter plots: running time vs. the values taken by the target hyperparameters concerning LSM-I-12Year.png

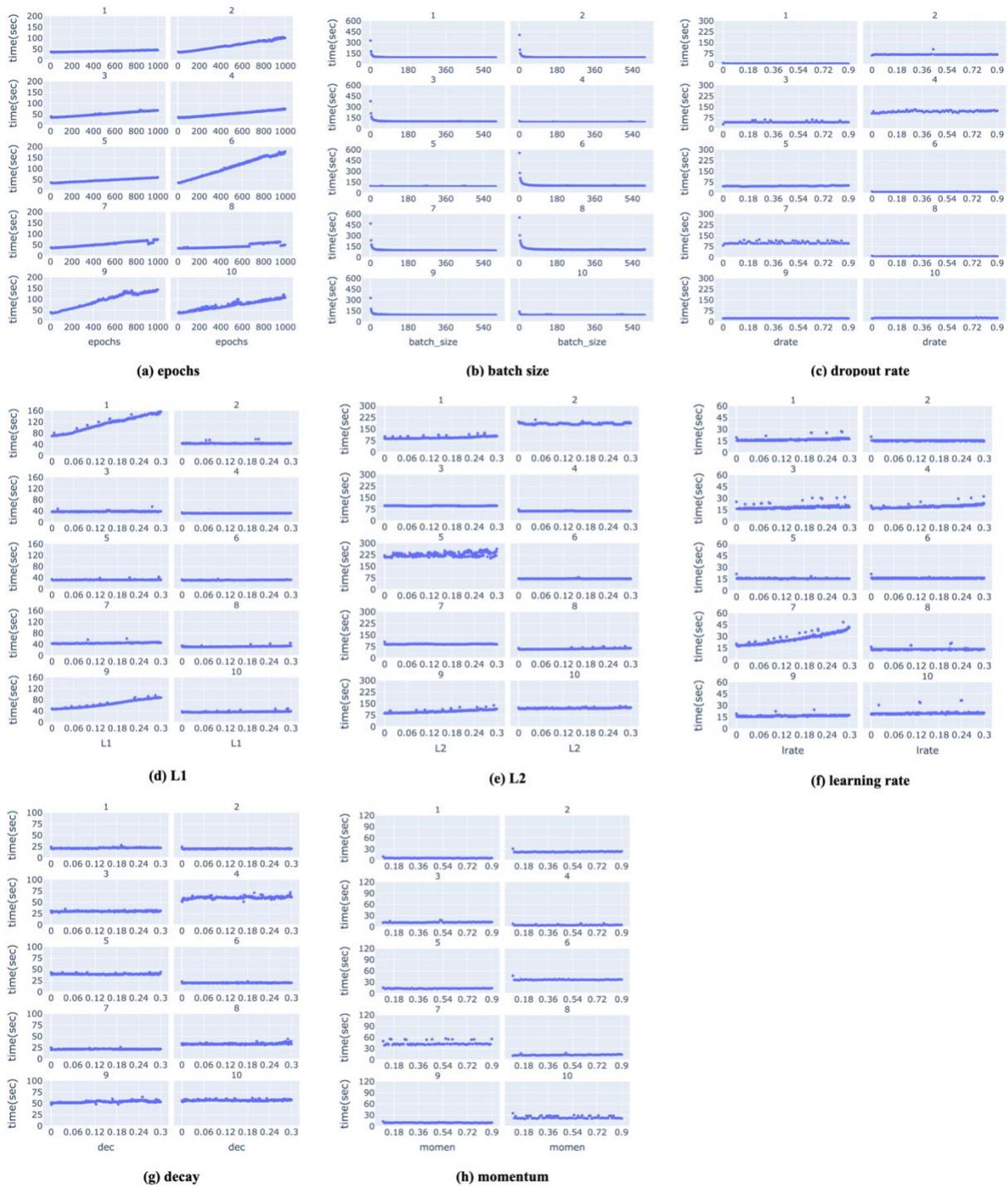

**Figure S3.** Scatter plots: running time vs. the values taken by the target hyperparameters concerning LSM-I-15Year